\documentclass{IOS-Book-Article}

\usepackage{mathptmx}
\usepackage[utf8]{inputenc}
\usepackage{hyperref}
\def\hb{\hbox to 10.7 cm{}}

\newcommand{\synset}[2]{{\{\texttt{#1}: \textit{#2}\}}}

\begin{document}

\pagestyle{headings}
\def\thepage{}

\begin{frontmatter}
\title{Linguistic Legal Concept Extraction in Portuguese}
\author[A]{Alessandra Cid}
\author[B]{Alexandre Rademaker}
\author[C]{Bruno Cuconato}
\author[D]{Valeria de Paiva}

\address[A]{FGV/Direito Rio and FGV/EMAp}
\address[B]{IBM Research and FGV/EMAp}
\address[C]{FGV/EMAp}
\address[D]{Nuance Communications}

\begin{abstract}
This work investigates legal concepts and their expression in Portuguese, concentrating on the ``Order of Attorneys of Brazil'' (in Portuguese `Ordem dos Advogados do Brasil' or OAB) Bar exam and especially on its questions on Ethics, a subset of the multiple questions this national bar exam deals with. Using the corpus formed by a collection of multiple-choice exams, three norms related to the Ethics part of the OAB exam (the law 8906, OAB Ethics Code and OAB's General Regulation of the law 8906) and some language resources (Princeton WordNet and OpenWordNet-PT) and tools (AntConc and Freeling), we began to investigate the concepts missing from our repertory of concepts and words in Portuguese, as expounded in our basic lexical resource, the knowledge base OpenWordNet-PT. We add these concepts to a special glossary and hence obtain a representation of these texts that is mostly  ``contained'' in the lexical knowledge base. These eventually can be used to calculate entailment and contradiction between passages, which would support the answer the questions automatically.
\end{abstract}

\end{frontmatter}

\section{Introduction}

Becoming a lawyer is a widely varied process around the world. Common to all jurisdictions are requirements of age and competence; some jurisdictions also require documentation of citizenship or immigration status. However, the most varied requirements are those surrounding the obtaining of the license to practice, whether it includes finishing a law degree, passing an exam, or serving an apprenticeship. Basic requirements vary from country to country.

In Brazil, the ``Order of Attorneys of Brazil'' (in Portuguese `Ordem dos Advogados do Brasil' or OAB), the Brazilian Bar association, administers a bar examination nationwide three times a year. The exam is divided in two stages -- the first consists of 80 multiple choice questions covering several disciplines (e.g. Ethics, Human Rights, Philosophy of Law, Constitutional Law, Administrative Law, Civil Law, etc). The candidate must score at least 40 questions correctly to proceed to the second part of the exam, which requires answering four essay questions and a drafting project (writing a motion, opinion or claim document). Success in the examination allows one to practice in any court or jurisdiction of the country. However only 20\% of those who take the exam are successful.

Our working hypothesis is that to answer correctly the multiple choice questions of the OAB Bar examination, candidates need to construct a model of the world, which consists of concepts that they learn when reading the text of the laws and jurisprudence associated with the questions. We would like to develop a computer system capable of mimicking some of these human abilities, using Natural Language Processing (NLP) tools.

In the next section, we describe some NLP tools, including our main lexical resource, the OpenWordNet-PT (abbreviated as OpenWN-PT or simply OWN-PT). Then we discuss extensions to deal with legal domains and the methodology we developed to deal with this kind of specific, very sophisticated domain. 

\section{The OAB Bar Examination}

The OAB Bar examination provides a sensible benchmark to evaluate a system attempting to provide question-answering facilities, based on Brazilian laws and regulations. An ideal legal question-answering system would take a question \texttt{Q} in natural language and a corpus of all legal documents in a given jurisdiction \texttt{LawCorpus}, and would return both a correct answer (easier if using multiple choice) and its legal foundation, i.e., which sections of which norms or laws provide support for the answer and why this is so. As stated this is too broad and too hard: we hope to provide a sample corpus (a subset of \texttt{LawCorpus}) with a single detailed law, to see how far we can get the processing to go. We then go on to describe a methodology for continuously improving our processing of the vocabulary of the law.

Previous work on the corpus constructed from multiple choice questions attests to the suitability of the data obtained from the OAB Bar questions. The data from OAB's previous exams and their answer keys were obtained as PDF files from the official source at \url{http://oab.fgv.br/}, cleaned and prepared for processing with some scripts \cite{delfino2017}. In \cite{delfino2017} we also described a simple question answering system targeting the exams, based on shallow NLP methods. In \cite{delfino2018} we improved the system by incorporating wordnet data to its analysis process, and started doing a very preliminary effort of expanding OWN-PT to the legal domain. However, it is clear from inspection (and previous work) that the Law domain has many concepts and words that are only used within the legal profession. If they are to be used to reason about the Law, these concepts and words need to be added to OWN-PT, our basic lexical resource, described below. This work follows this effort, so that future systems can benefit from a legal lexical semantic database.

\section{OpenWordNet-PT}

The OpenWordnet-PT \cite{depaiva2012} is an open access wordnet for Portuguese, originally developed by Valeria de Paiva, Alexandre Rademaker and Gerard de Melo as a syntactic projection of Universal WordNet (UWN) \cite{uwn}. When the project on OWN-PT started there was no open wordnet-like resource for Portuguese.

The process of building the OWN-PT used machine learning to construct relationships between graphs representing information coming from several versions of Wikipedia, as well as from open dictionaries. Starting as a projection at the level of the lemmas in Portuguese and their relationships, the OWN-PT has been constantly improved through linguistically motivated additions, manual and semi-automatic, making use of large corpora. This kind of construction, automatically started, but manually curated and improved, is the hallmark of our work using OWN-PT.

The OWN-PT has been developed since 2010 with the main objective of eventually serving as a lexicon for a proposed NLP system focused on logical reasoning, based on knowledge representations coming from language.
The philosophy of OWN-PT is to maintain a close alignment with the original Princeton WordNet (PWN) \cite{wordnet1998}, but to remove the biggest mistakes created by automated methods, using language tools and skills for this cleaning up task. One positive consequence of the close connection between PWN and OWN-PT is the latter ability to minimize the impact of lexicographical decisions on the separation or grouping of senses in a synset, as these decisions are, for the time being, delegated to PWN, to a large extent. We strive for precision, rather than coverage, as far as OWN-PT is concerned, and precision is surely needed when dealing with the legal domain. 

\section{Analyzing the legal vocabulary}

We want to make sure that the basis created with OWN-PT is broad enough to allow us to deal with specific domains such as Law or Geology. Clearly these specific domains have specific vocabulary, both in terms of words that are not part of the everyday vocabulary, but specially in the use of expressions. In Law, there are several expressions that are not, as yet, part of the OWN-PT. Many common nouns are missing. A significant number of these are nominalizations such as \textit{impetração} (a kind of filing), \textit{postulação} (postulation), where the verbs \textit{impetrar} and \textit{postular} (to file, to postulate) are already in the OWN-PT lexicon.

Some of the missing expressions are adjectives, like \textit{fundacional} (foundational) and \textit{constitutivo} (constituent) coming from nouns \textit{fundação} (foundation) and \textit{constituição} (constitution), where sometimes PWN prefers not to list all the morphologically derived expressions. Still others are nouns that are nominalizations of adjectives, like \textit{nulidade} (nullity), derived from \textit{nulo} (null), where, again, morphology could play am important role. 

Expressions such as the name of a law, e.g. \textit{Estatuto da Advocacia} or the name of the professional association of lawyers in Brazil, the ``Ordem dos Advogados do Brasil'' (OAB) are essential synsets that needed to be created. These are expected, since the named Brazilian legal entities are clearly different from the American ones. We need synsets corresponding to the ones for \textit{President of the United States} and \textit{U.S. Congress}, for instance. 

There are also other, more general legal expressions, called \textit{multiword expressions} (MWEs), that really describe the field, but are harder to deal with. Some of these are in Latin, such as \textit{habeas corpus}
or \textit{data venia}. But most others are simply common Portuguese words, used in fixed expressions, which have more specific meanings. For example the expression \textit{defensor público}
could be used for someone who defends the public or someone who defends something in public, but it is mostly used to describe the attorney, appointed by the Estate to defend the interests of poor citizens, who are not able to pay for a lawyer. 

Some recent work on these multiword expressions, especially on English noun compounds \cite{farahmand2015multiword,Sag:2002}, makes the point that multiword expressions can be compositional or non-compositional, conventionalized and not conventionalized. In \cite{farahmand2015multiword}, it is said in the introduction: 

\begin{quote}
The lack of practical data sets that can be used in the training and evaluation of multiword expression (MWE) related systems is a notorious problem \cite{mccarthy2003detecting,hermann2012unsupervised}. It is partly due to the heterogeneous nature of MWEs, partly due to their frequency, and partly due to the unclear boundaries between MWEs and regular phrases. These issues have made the compilation of useful MWE data sets challenging, and any effort to create them invaluable.
\end{quote}

It is abundantly clear that specific domains like Law require a bigger set of MWEs, both compositional (or not) and conventionalized (or not). Briefly we can say that \textit{semantic non-compositionality} is the property of a compound whose meaning can not be readily interpreted from the meanings of its components. \textit{Conventionalization} refers to the situation where a sequence of words that refer to a particular concept is commonly accepted in such a way that its constituents cannot be easily substituted for near-synonyms, because of some cultural or historical conventions. A large fraction of compounds are to some extent conventionalized, however we are interested in only clear and well-known conventionalizations, which \cite{farahmand2015multiword} refer to as ``marked conventionalization''. We assume that non-compositional compounds are by definition conventionalized, hence it only makes sense to consider conventionalization (or not) of compositional compounds. 

It is also clear that non-compositional MWEs are easier to spot: for example \textit{má fé} (bad faith) has nothing to do with \textit{fé} (faith) in its most used meaning of `religious belief'. It simply means ``in a deceiving way'' and it is not specific to Law, but common currency both in Portuguese and in English, where the synset \synset{00753240-n}{for double-dealing, duplicity} has a gloss which reads \textit{acting in bad faith; deception by pretending to entertain one set of intentions while acting under the influence of another}. 

\section{Experiments}

In order to identify relevant legal terms and multiword expressions and to analyze how this legal vocabulary can be incorporated to the OWN-PT, we describe three small experiments with legal texts and the OWN-PT. 

In our first experiment, we investigated some of the English terms in the PWN synsets that were classified by the synset\footnote{A wordnet is a lexical database that groups nouns, verbs, adjectives and adverbs into sets of cognitive synonyms (synsets), each expressing a distinct concept. Synsets are interlinked by means of conceptual-semantic and lexical relations.} \synset{08441203-n}{jurisprudence, law} in PWN/OWN-PT.\footnote{PWN contains many semantic relations between synsets, besides the most well-known \texttt{hyponym}, \texttt{hypernym}, and \texttt{antonym}, we also have the relation \texttt{classifiedByTopic} for grouping synsets into domains.} Our hypothesis was that, by translating the English terms that were already classified as legal vocabulary, we would incorporate important legal terms in Portuguese to the OWN-PT. We analyzed some of the terms that were on the topic, verifying the quality of the translations of the terms that were already translated and seeing if we could translate the ones that were not. We reached the conclusion that the synsets related to \synset{08441203-n}{jurisprudence, law} were very specific to American Law and that by adding their translations to the OWN-PT we were not expanding it with relevant words for legal vocabulary in Portuguese. For example, we found several expressions for specific types of laws in English, such as ``Gag Law'' or ``Blue sky law'', that are not used in Portuguese. Given this situation, we moved on to our second experiment. 

Our second experiment deals with a wholesale construction of a glossary of legal terms extracted from the OAB questions and from the three norms that are the base of the Ethics questions of the OAB exam. These last three norms are: the law 8906 of July of 1994, the `Código de Ética da OAB' (Ethics Code of the OAB) and the `Regulamento Geral da OAB' (OAB's General Regulation). We analyzed these documents using  AntConc \cite{antconc}, a corpus analysis toolkit for concordancing and text analysis. Using AntConc, we obtained a list of 6,890 bi-grams and tri-grams on the texts that occur more than 9 times. Since AntConc works over the raw text, without using any linguistic annotation, we had to filter n-grams that were clearly not MWEs. Two annotators filtered the list independently and we combined the results ending up with 430 candidates of MWEs.

Following \cite{farahmand2015multiword}, instead of deciding which n-grams are true MWEs as opposed to simple collections of words that happen to occur together, we used a simple test to classify each candidate as compositional or non-compositional and conventional or non-conventional. Is the meaning of this expression explained by the meanings of its parts? If not, then we think we have a non-compositional MWE. If the meaning of the expression is compositional, is it a title of an article in the Portuguese Wikipedia?\footnote{We obtained a list of titles of all Portuguese pages from Wikipedia at \url{https://dumps.wikimedia.org/other/}.} If yes, we reckon this is sufficient evidence to characterize a conventional MWE. 
If it is not a Wikipedia title, it may be that Wikipedia should have one such page and is missing it. Therefore, our process is probably an oversimplification that could be improved in the future. 
Finally, for the top 50 most frequent MWEs, we identified the possible head words from expressions. We assigned the nouns to their respective synset on OWN-PT and the expressions to their hyponym.\footnote{The list of MWEs and all data from the experiments will be made available upon acceptance.}


In our third experiment, we investigated the lexical units of the law 8906, one of the norms used in the second experiment. Law 8906 describes the rights and obligations of lawyers and how they can advocate for their clients. Since the Ethics part of the Bar examination is one of the most straightforward sections of the exam, it makes sense for us to make sure that the whole law is processed correctly and that all the required vocabulary is in place, before starting to analyze the law itself and before trying to relate the OAB ethics questions to their answers.

The experiment was carried out using Freeling~\cite{padro2012}, an NLP library able to process Brazilian Portuguese. We processed the Law 8906, investigating the results of the tokenization, lemmatization, part-of-speech (PoS) tagging and word sense disambiguation, that is, if they were correctly assigned to OWN-PT senses in the context of the articles of the law. This allowed us to evaluate how Freeling's modules could be adapted to process the law more accurately and enabled us to measure how many words belonging to the legal vocabulary were already on OWN-PT or needed to be added. 

Some of Freeling's results after processing the law were expected. Since OWN-PT, just as Princeton WordNet, does not cater for pronouns, determiners or prepositions, it did not have a meaning assignment for these cases. Freeling lemmatization and PoS tagging modules are driven by a dictionary of word forms. The words that are not in Freeling's dictionary must have the lemma and part-of-speech tag guessed which introduce some errors. For example, the Portuguese word \textit{juizado} (court) was not in the dictionary, so its lemmatization was wrongly ascribed as \textit{juizados}. This was evidence that FreeLing's dictionary did not have it and we simply added it. 
The multiword expressions identified and added to OWN-PT must also be added to the Freeling locutions file, so that tokens that are part of an MWE are joined enabling the word sense disambiguation module to associate the whole expression to an OWN-PT synset. Other bugs are still under investigation, but the results we obtained so far are summarized in Table~\ref{tab:freeling}, where we present basic statistics of Freeling's analysis of Law 8906. To obtain the unique totals we considered the pair (lemma,PoS tag), and we only considered that a word was missing a sense if it was tagged as the right PoS. Law 8906 comprise 87 articles summing up 231 sentences and 10,242 tokens (1,508 unique types/words). Table~\ref{tab:freeling} shows in the last column that we are still missing some words in OWN-PT. 

As a way of completing OWN-PT, one can discover easy synsets that needed completion. 
For example, the synsets \synset{01987341-a}{reserved (marked by self-restraint and reticence; ``was habitually reserved in speech, withholding her opinion'')} and \textit{01988324-a}{reserved (set aside for the use of a particular person or party)}, are very easy to complete. The word \textit{reservado} is almost exactly the same as the English one and has the associated adverb \textit{reservadamente} corresponding to the also empty synset \synset{00441649-r}{reservedly (with reserve; in a reserved manner)}. There are plenty of occasions where English and Portuguese conceptualize things differently and these require plenty of effort. Finding the easy synsets, where not only concepts, but even words are almost the same and making sure that these are complete in the lexical base is one of our main goals for OWN-PT.

\begin{table}
\begin{center}
\begin{tabular}{lrrr}
& total & unique & no sense \\
\hline
Nouns & 2629 & 727 & 190\\
Adjectives & 634 & 234 & 60\\
Verbs & 1167 & 330 & 16\\
Adverbs & 268 & 77 & 32\\
\end{tabular}
\end{center}
\caption{Analysis of Law 8906 by Freeling}\label{tab:freeling}
\end{table}



\section{Conclusions}

This preliminary work investigates legal concepts and their expression in Portuguese, concentrating on the OAB Bar exam and on one of the Ethics' laws, that a subset of the exam questions are based on. Using the corpus formed by the collection of multiple-choice questions in the exams, this one specific law and some language resources (PWN and OWN-PT) and NLP tools (AntConc and Freeling) we began to investigate the concepts missing from our repertory of concepts and words in Portuguese, the knowledge base OWN-PT.

This initial work does not require a huge effort of evaluation, since we are mostly discovering word forms and senses that the lexical resource does not have, yet. We can count how many word forms were added to synsets, how many MWEs we had to create, but we have not found baselines to compare our work to, so far.

As maintainers and curators of OWN-PT this kind of work provides a way of dicing up a nice subset of synsets to look up and correct. Most of the work we have done so far on improving OWN-PT has been based on grammatical functions: we had a push to improve the verb lexicon \cite{verb-lexicon2016}, another push to provide nominalizations and their links \cite{nomlex2014}, an effort to increase demonyms and gentilics \cite{gentilics2016}, which was meant to break down the huge class of adjectives into smaller subsets. We have not done as much in terms of topics or semantic domains. We did a preliminary study of Geological Eras \cite{gwn2018}, but this was mostly to check the feasibility of merging an external ontology to the subjacent hierarchy of OWN-PT noun synsets. We also did a preliminary assessment of temporal related concepts \cite{lrec2018temporal}, using HeidelTime \cite{heideltime2013}. But this is our first time tackling a sophisticated and specialized field like Law. 

We are aware of several difficulties ahead. Firstly the judicial system in Brazil (based on Roman law) is very different from the `Common Law' system in use in the US and UK, where most of the lexical resources we want to make use of, originate. This difficulty is discussed in \cite{bertoldi2011}, who wanted to use FrameNet \cite{framenet2003} to create a corpus encompassing the whole judicial system in Brazil. They say that ``This[their proposed] corpus is being planned to be representative of the entire legal production of Brazilian courts and Brazilian legislative houses, such as the laws published by the Brazilian Senate and the judicial decisions of the Federal Courts''. We are less ambitious, we hope to produce a corpus of laws and regulations that allow us to answer the Ethics questions of our collection of OAB exams, at least to begin with. 

Secondly we would like to produce a large glossary of legal terms that could be used for students actively taking the OAB exam, focusing on the multiple choice questions. There are several good juridical dictionaries in Portuguese and in English, but it seems to us that an open-source one, with relations of synonymy and antonymy, mostly based on the past OAB entrance examinations would be useful to students and professors of Law. 
Thirdly, we reckon that this project, although ambitious, would allow us to push on the direction we really want to work on, that is, the direction of reasoning with the contents of the legal texts. We plan to start using shallow methods, but also want to try our hand at deep logical representations, hybridized together with learning approaches, to try and detect entailment and contradiction between pieces of legal text.

As for future work, we need to complete the glossaries that we started constructing. When the mappings are consistently investigated, we need to establish a process to make sure that newer changes do not undermine the previous work, i.e. we need to establish test suites and regression tests, as described for instance in \cite{depaiva2008}. We also need to expand and complete our glossary of juridical terms, making sure that they fit in with what is provided by open source juridic dictionaries, such as the Legal Dictionary of `The Free Dictionary by Farlex'.\footnote{\url{https://legal-dictionary.thefreedictionary.com/}} Finally we would like also to design and implement our own system for computing ``entailment and contradiction detection'' between the OAB examination questions and their answers.

\bibliography{references}
\bibliographystyle{plain}

\end{document}